\documentclass[12pt,]{article}
\usepackage{amsmath}
\usepackage{amssymb}
\usepackage{geometry}
\begin{document}
\title{Note on the Construction of Structure Tensor}
\author{Josef Bigun \and Fernando Alonso-Fernandez}
\date{}
\maketitle
\begin{center}
Halmstad University\\
30117 Halmstad, Sweden\\
\texttt{josef.bigun@hh.se}, \texttt{fernando.alonso-fernandez@hh.se}
\end{center}

\begin{abstract}
This note presents a theoretical discussion of two structure tensor constructions: one proposed by Bigun and Granlund 1987, and the other by Granlund and Knutsson 1995. At first glance, these approaches may appear quite different--the former is implemented by averaging outer products of gradient filter responses, while the latter constructs the tensor from weighted outer products of tune-in frequency vectors of quadrature filters. We argue that when both constructions are viewed through the common lens of Total Least Squares (TLS) line fitting to the power spectrum, they can be reconciled to a large extent, and additional benefits emerge. From this perspective, the correction term introduced in Granlund and Knutsson 1995 becomes unnecessary. Omitting it ensures that the resulting tensor remains positive semi-definite, thereby simplifying the interpretation of its eigenvalues. Furthermore, this interpretation allows fitting more than a single 0rientation to the input by reinterpreting quadrature filter responses without relying on a structure tensor. It also removes the constraint that responses must originate strictly from quadrature filters, allowing the use of alternative filter types and non-angular tessellations. These alternatives include Gabor filters--which, although not strictly quadrature, are still suitable for structure tensor construction--even when they tessellate the spectrum in a Cartesian fashion, provided they are sufficiently concentrated.
\end{abstract}
 
\subsection*{1. Introduction}

The Structure Tensor for N-dimensional images was introduced in \cite{bigun87london} using matrix eigenanalysis. It was also shown to be equivalent to complex convolutions and pixelwise complex squaring in 2D images, without relying on matrix algebra. Simultaneously and independently, the 2D Structure Tensor in matrix form was introduced by two other groups \cite{kass}, \cite{forstner87}.

The tensor has been widely used in directional analysis of images, often relying on derivatives of Gaussians for filtering. It is also used to analyze responses from direction- and frequency-tuned quadrature filter banks. In the latter case, the tensor enables interpolation of sharp directions, as introduced for 3D images in \cite{knutsson89tensor}, replacing earlier approaches based on 5D mappings \cite{knutsson1985producing}. Thus, the Structure Tensor can be implemented either directly via Gaussian derivatives or by applying it to the magnitude responses of quadrature or Gabor filters.
It has proven to be a robust tool for performing Total Least Squares (TLS) estimation of a single orientation model in local image regions. For this reason, it is justified to discuss common constructions of the tensor.

For readers interested in fitting multiple orientations to the power spectrum—especially in 2D—we refer to the multi-orientation Structure Tensor \cite{bigun94pami2}, which extends the original concept. This extension uses higher-order spectral moments, implemented via Gabor filter responses.

To avoid confusion, we note that the present discussion does not address another extension of the Structure Tensor that models orientation using harmonic coordinates. This extension is particularly suited for recognizing objects with unique and precise reference or singularity points—such as circles, cross junctions, or parabolas \cite{bigun88amst}. It also relies on higher-order complex derivative filters, but applies them in a different order relative to the nonlinearity introduced by complex squaring.

Filters with transfer functions of the form $(\omega_x + i \omega_y)^n G(\omega_x, \omega_y)$, where $G$ is a Gaussian, are used in all three versions of the Structure Tensor in 2D by varying $n$ and the placement of the filter in the processing chain. These filters are also related in form to linear rotation-invariant filters \cite{danielsson80} and steerable filters \cite{freeman91}. However, the Structure Tensor and its extensions in 2D are not linear, as the filtering results are intertwined with complex squaring. The remarkable properties of these complex filters—especially when cascaded or analyzed via the Fourier Transform—are detailed in \cite{bigun04pami3}.

More recently, signal representations involving phase, orientation,
absolute frequency, and magnitude extraction have employed the Riesz
transform, which uses analogues of these complex filters in 2D \cite{larkin}, \cite{Felsberg01sp}, \cite{puspoki16}. The Structure Tensor, constructed via gradients or quadrature filters, has also proven useful for TLS-based orientation estimation required by such signal representations.

In N-D, the Riesz transform is linear and has the transfer function $\boldsymbol{\omega}/|\boldsymbol{\omega}|$, corresponding to a smoothed image that is additionally filtered by $\boldsymbol{\omega}$—a component also used in Structure Tensor constructions. This further motivates a discussion of different constructions of the Structure Tensor, particularly those using quadrature filters.

Two such constructions will be detailed for  Structure Tensor
in its original matrix formulation. Both are intended to complement
spectral realizations—such as quadrature or Gabor filters—for
evaluating the presence of a single sharp orientation, for 2-D and
higher dimensional images.

\subsubsection*{2. Structure Tensor construction by enforcing Linear Symmetry in the input} 

In \cite{Granlund95,Knutsson09}, Granlund and Knutsson assumed a locally
linear symmetric image\footnote{The term is defined in \cite{bigun87london} as  
$ f(\mathbf{r}) = g(\mathbf{k}^\top \mathbf{r}) $ where $g(\tau)$
is a scalar function of one variable, $\tau$, and $\mathbf{k}$ is a normalized
vector. Knutsson and Granlund use the term ``simple neighborhood'' for
this type of input. In the literature, {\em intrinsically 1D} is also used for the same concept.} that excites quadrature filters. These filters tessellate half of the power spectrum systematically, albeit with overlaps. The structure tensor that would result without overlaps, when the assumption holds, can be reconstructed, among other methods, by applying frame theory techniques to the {\em magnitude responses } $q_k \ge 0$. It results in a Structure Tensor whose diagonal elements must be corrected by an additional term.

A structure tensor is thus computed for each local image region using
$q_k$, which serve as weights for frequency coordinate tensors based
on the filters’ tune-in direction vectors $\mathbf{n}_k \in
\mathbb{R}^N$. For example, for a 3D Structure Tensor and icosahedral tessellation with six filters, this leads to the following estimation comprising the negative correction term:
\begin{equation}
\mathbf{T}_{GK} = \sum_{k=1}^{6} q_k \left( \frac{5}{4} \mathbf{n}_k
\mathbf{n}_k^\top - \frac{1}{4} \mathbf{I} \right), \quad
\text{where } \mathbf{I} \text{ is the identity matrix.}
\label{eq:Granlund09}
\end{equation}

The six direction vectors $\mathbf{n}_k$ are:
\[
\begin{aligned}
\mathbf{n}_1 &= (a, 0, b), & \mathbf{n}_2 &= (-a, 0, b), \\
\mathbf{n}_3 &= (b, a, 0), & \mathbf{n}_4 &= (-b, a, 0), \\
\mathbf{n}_5 &= (0, b, a), & \mathbf{n}_6 &= (0, -b, a),
\end{aligned}
\]
where
\[
a = \frac{2}{\sqrt{10 + 2\sqrt{5}}} \approx 0.525731, \quad
b = \frac{1 + \sqrt{5}}{\sqrt{10 + 2\sqrt{5}}} \approx 0.850651.
\]

The above construction follows the descriptions in
\cite{Granlund95,Knutsson09}, though they are not fully explicit in motivation. Notably, the same interpretation has been independently made by Felsberg in \cite{Felsberg02phd}, via eq.~3.49 but for the 2D case.

However, the resulting $\mathbf{T}_{GK}$ may not be positive semi-definite due to the negative term in eq.~(\ref{eq:Granlund09}). In the example, if one happens to obtain filter response magnitudes:
\[
q_1 = 1, \quad q_3 = q_5 = 0.25, \quad q_2 = q_4 = q_6 = 0,
\]
this yields a $\mathbf{T}_{GK}$ with eigenvalues
\[
\lambda_1 = \lambda_2 = -0.3750, \quad \lambda_3 = 2.8975,
\]
indicating that two of the eigenvalues are negative. As discussed further below, the indefiniteness may be problematic in applications, especially those relying on the assumption that eigenvalues model errors, which can never be negative.

\subsubsection*{3. Structure Tensor Construction by Direct Sampling of Required Functions}

A structure tensor, on the other hand, is always symmetric and
positive semi-definite, even when derived from an arbitrary local image (not just
from linearly symmetric inputs, also called intrinsically 2D). However, this requires constructing and interpreting
the structure tensor in terms of Total Least Squares (TLS) line
fitting directly to the power spectrum. This interpretation does not require that the implementation be done via gradient filtering.

The tensor
$\mathbf{T}_{BG}$, in its initial definition as a weighted sum of outer products of spectral coordinates—which for implementation purposes was translated to a sum of outer products of gradients in the spatial domain—can achieve this. 
\begin{equation}
\mathbf{T}_{BG} = \sum_{k=1}^{6} q_k \mathbf{n}_k \mathbf{n}_k^\top. \label{eq:bigunst} 
\end{equation}
The spectral-domain formulation of the gradient-based construction is thus retained here for consistency with $\mathbf{T}_{GK}$, which is also defined in the spectral domain \cite{bigun87london}.

This construction assumes that filter magnitude responses 
directly represent or sample the power spectrum magnitudes (i.e., squared magnitudes). This assumption is justifiable from a signal-theoretic perspective, as taking scalar products of interpolator functions with the function to be sampled is a valid sampling method. The fact that the sampled function resides in the spectral domain does not alter this fundamental principle. Whether the interpolating filters have large or small angular overlaps is important, but is a design choice depending on the application. For instance, if a coarse model of image orientation suffices, large angular overlaps—as provided by default quadrature filters of \cite{Granlund95}—are appropriate. Conversely, for fine-grained orientation analysis, filters with narrower angular support, such as Gabor filters or sharpened quadrature filters with minimal overlap, are more suitable.

The idea that tensor elements inherit the sampling of the spectrum—i.e., by directly multiplying frequency coordinates without the interpolation implied by the frame-based construction in \cite{Granlund95,Knutsson09}—is also justifiable. The tensor elements represent coordinate-weighted spectral energies, which are bandlimited if the input image is bandlimited, although the bandwidth may double in the worst case. We elaborate on this point below.

Assuming the input is $f(\mathbf{r})$, the second-order spectral energy moments can be computed either in the frequency domain by summing
\begin{equation}
\|F(\mathbf{\omega})\|^2 \, \mathbf{\omega} \mathbf{\omega}^\top
\end{equation}
at discrete frequency coordinates $\mathbf{\omega}_k$, or in the spatial domain by summing
\begin{equation}
\nabla f(\mathbf{r}) \nabla^\top f(\mathbf{r})
\end{equation}
at discrete image points make both methods of construction equivalent. Sampling spectral energy is valid provided that the involved functions are bandlimited in the other domain, allowing the use of  tools such as  convolution, and pointwise multiplication via  Fourier series or DFT to implement convolution, or use Parseval-Plancherel theorem. In particular, the sampling frequency must exceed the Nyquist rate in the concerned spectral coordinate.

Multiplication of two functions, or squaring in one domain, is convolution in the other domain whereby the bandwidth increases in the latter domain. The bandwidth may thus increase by a factor of two compared to the original spectrum, but this is manageable. The logic follows from the fact that if $F$ can be sampled without loss, then it contains no components above $\pi$ (the Nyquist frequency) in any coordinate direction in the other domain. Consequently, $F$ can be reconstructed and upsampled, allowing the derivative cross-products or second-order spectral energies to be computed without information loss from the discrete  $F$. As long as the function to be sampled is  bandlimited, its square, magnitude square, derivatives, and mixed-products products of real and imaginary parts  will also be bandlimited and can be sampled directly. For example, the square of $F$ at an image point can be computed as the square of the sample at that point, without loss of information. 

It is worth noting that upsampling is only necessary if the function to be sampled contains significant content between half the Newquist ($\pi/2$)  and and the Newquist frequency\footnote{In practice, most images are sampled densely enough so that upsampling is not required. Moreover, derivative approximations include a low-pass component—i.e., they are derived from low-pass interpolants and then sampled—implicitly suppressing high-frequency content in the critical band  (between $\pi/2$ and $\pi$). While upsampling is not computationally expensive, avoiding it can still save resources. These are  main reasons for why upsampling is often omitted in implementations of structure tensor computation. However, these are also engineering considerations and do not negate the theoretical possibility of upsampling to avoid aliasing when need be in direct tensor sampling.} and the content is worth to keep for the application, i.e. not noise which can be suppressed via linear filters in the process. Although more involved, the inheritance of bandlimitedness from $F$ or $f$ to compute products of gradient components or corresponding coordinate products weighted with magnitudes (moments) can also be extended to sampling in polar coordinates.


A key advantage of constructing the structure tensor by directly sampling the required functions is that the resulting expression, as in (\ref{eq:bigunst}), contains no negative term. The tensor $\mathbf{T}_{BG}$ is therefore always positive semi-definite by construction. This expression measures second-order moments of the power spectrum, which are theoretically proven to enable TLS-based orientation estimation. In this framework, the eigenvalues of the tensor quantify modeling errors under the assumption that the input exhibits linear symmetry. If the Linear Symmetry model does not hold, the eigenvalues still serve as reliable indicators of the deviation, as they are guaranteed to be non-negative. The tensor eigenvectors correspond to directions optimally fit to the power spectrum, regardless of the input. When the Linear Symmetry model holds exactly—or when other types of linear symmetry in higher dimensions apply, such as concentration to $N-k$ dimensional hyperplanes in the spectrum \cite{bigun87london,jaehne97b}—this will be reflected in the eigenvalue distribution, including how many of them, and which ones, are close to zero. 

Another subtle but important distinction in the direct function sampling approach is the treatment of angular overlap in the frequency domain. The polar angular tessellation of the quadrature filters is designed to have maximal overlap by default. These filters decay to zero angularly at $\pm \pi/2$ from the tune-in orientation, following the cosine function, as required by the quadrature (Hilbert transform) condition. A slow angular decay implies strong low-pass filtering in angular spectral coordinates (orientation), meaning that a single quadrature filter responds to a broad range of orientations around its tune-in direction, rather than to a narrow band. This is acceptable when the filters are used solely to assess the presence of linear symmetry in the input, as  $\mathbf {T}_{BG}   $ will perform the necessary roll-back via the (nonlinear Karhunen-Loewe, or Principal Components) TLS estimation of orientation and the corresponding modeling error.

However, such broad angular support limits the ability to resolve multiple coexisting orientations. While the structure tensor can still be constructed in the same way, its evaluation of linear symmetry will only indicate that the model fails due to the presence of multiple orientations; it will not specify which orientation components are present. At this point, attempting to extract more information from the filter responses—when the structure tensor is uninformative as to which orientation components exist in the input—will also be ineffective, since each filter averages energy over a wide range of directions. To enable filter responses to reveal more about the orientation content, the angular overlap between filters must be reduced—for example, by using Gabor filters or by modifying quadrature filters to have sharper angular selectivity (e.g., by raising the cosine function to a higher power). In such cases, the filter responses can provide more detailed information about the distribution of orientations. The degree of overlap, as already indicated,  should be chosen based on the angular sampling theorem and the desire as to how many jointly occurring orienations will be evaluated (the specific questions being addressed in the application), in addition to what the structure tensor can reveal.

In contrast, the tensor $\mathbf{T}_{GK}$ does not perform TLS-optimal line fitting to the spectrum. Instead, it is constructed by linear interpolation from an ideal tensor under the assumption that local linear symmetry excites a specific set of filters. The correction term and weights depend on the geometry and overlap of the filter tessellation. Reducing overlap decreases the magnitude of the negative correction term. Although approximating $\mathbf{T}_{GK}$ post hoc by a rank-1 tensor has been proposed, the presence of negative eigenvalues complicates interpretation. This issue is especially problematic in higher-dimensional settings ($N > 2$), where an indefinite tensor may fail to reveal the presence of $N-k$ dimensional hyperplanes in the spectrum as plausible explanations for the input.

\subsubsection*{3. Conclusion}
When $\mathbf{T}_{GK}$ is \textbf{indefinite}, negative eigenvalues may interfere with feature detection—such as corners or phase estimation—which rely on eigenvalue interpretation. This issue does not stem from limitations of the quadrature filters, but from the tensor construction itself, which permits indefiniteness. The problem can be remedied by constructing tensors that fit hyper-lines to the power spectrum in the TLS error sense, rather than forcing their responses to approximate those of a linearly symmetric input.
\bibliographystyle{plain}
\bibliography{bibl}

\end{document}